\documentclass[10pt,twocolumn,letterpaper]{article}

\usepackage{cvpr}              

\usepackage{microtype}
\usepackage{graphicx}
\usepackage{booktabs,makecell} 

\usepackage[numbers]{natbib}
\usepackage[ruled,boxed,algo2e]{algorithm2e}
\usepackage{bm}
\usepackage{amsmath}
\usepackage{amssymb}
\usepackage{graphicx}
\usepackage{xcolor}
\usepackage{subcaption}
\usepackage{multicol}
\usepackage{multirow}
\definecolor{citecolor}{HTML}{0071bc}
\usepackage[pagebackref,breaklinks=true,letterpaper=true,colorlinks,citecolor=citecolor,bookmarks=false]{hyperref}
\usepackage[title]{appendix}

\def\cvprPaperID{3364}
\def\confName{CVPR}
\def\confYear{2022}

\newcommand{\reffig}[1]{Figure~\ref{fig:#1}}

\newcommand{\refsec}[1]{Section~\ref{sec:#1}}

\newcommand{\reftbl}[1]{Table~\ref{Tbl:#1}}

\newcommand{\lblfig}[1]{\label{fig:#1}}
\newcommand{\lblsec}[1]{\label{sec:#1}}
\newcommand{\lbleq}[1]{\label{eq:#1}}

\newcommand{\lbltbl}[1]{\label{Tbl:#1}}

\author{
Dian Chen
\qquad
Philipp Kr\"ahenb\"uhl
\\
UT Austin
}

\begin{document}

\title{Learning from All Vehicles}  

\maketitle

\begin{abstract}

In this paper, we present a system to train driving policies from experiences collected not just from the ego-vehicle, but all vehicles that it observes.
This system uses the behaviors of other agents to create more diverse driving scenarios without collecting additional data.
The main difficulty in learning from other vehicles is that there is no sensor information.
We use a set of supervisory tasks to learn an intermediate representation that is invariant to the viewpoint of the controlling vehicle. 
This not only provides a richer signal at training time but also allows more complex reasoning during inference.
Learning how all vehicles drive helps predict their behavior at test time and can avoid collisions. 
We evaluate this system in closed-loop driving simulations. 
Our system outperforms all prior methods on the public CARLA Leaderboard by a wide margin, improving driving score by 25 and route completion rate by 24 points.

\end{abstract}

\section{Introduction}

Autonomous driving has been one of the most anticipated technologies since the advent of modern-day artificial intelligence.
However, even after decades of exploration, we have yet to see self-driving cars deployed at scale.
One main reason is the generalization.
The world and its drivers are more diverse than current planning approaches can handle.
Hand-designed classical planning~\cite{urmson2008autonomous,leonard2008perception,dolgov2008practical,bacha2008odin} does not generalize gracefully to unseen or unfamiliar scenarios.
Learning based methods~\cite{pomerleau1989alvinn,chen2015deepdriving,codevilla2018end,bansal2018chauffeurnet,chen2019lbc} fare better, but suffer from a long tail of driving scenarios.
The majority of driving data consist of easy and uninteresting behaviors.
After all, humans drive thousands of hours before observing a traffic accident~\cite{tefft2017rates}, especially when driving an expensive autonomous test vehicle.
How do we tame the long-tail of driving scenes?
While many approaches rely on carefully crafted safety-critical scenarios in simulation~\cite{suo2021trafficsim,lopez2018microscopic,peng2021learning}, or collect massive data in the real world~\cite{bansal2018chauffeurnet,sun2020scalability}, in this paper we focus on an orthogonal direction.

\begin{figure}[t]
\centering
\includegraphics[width=0.95\linewidth]{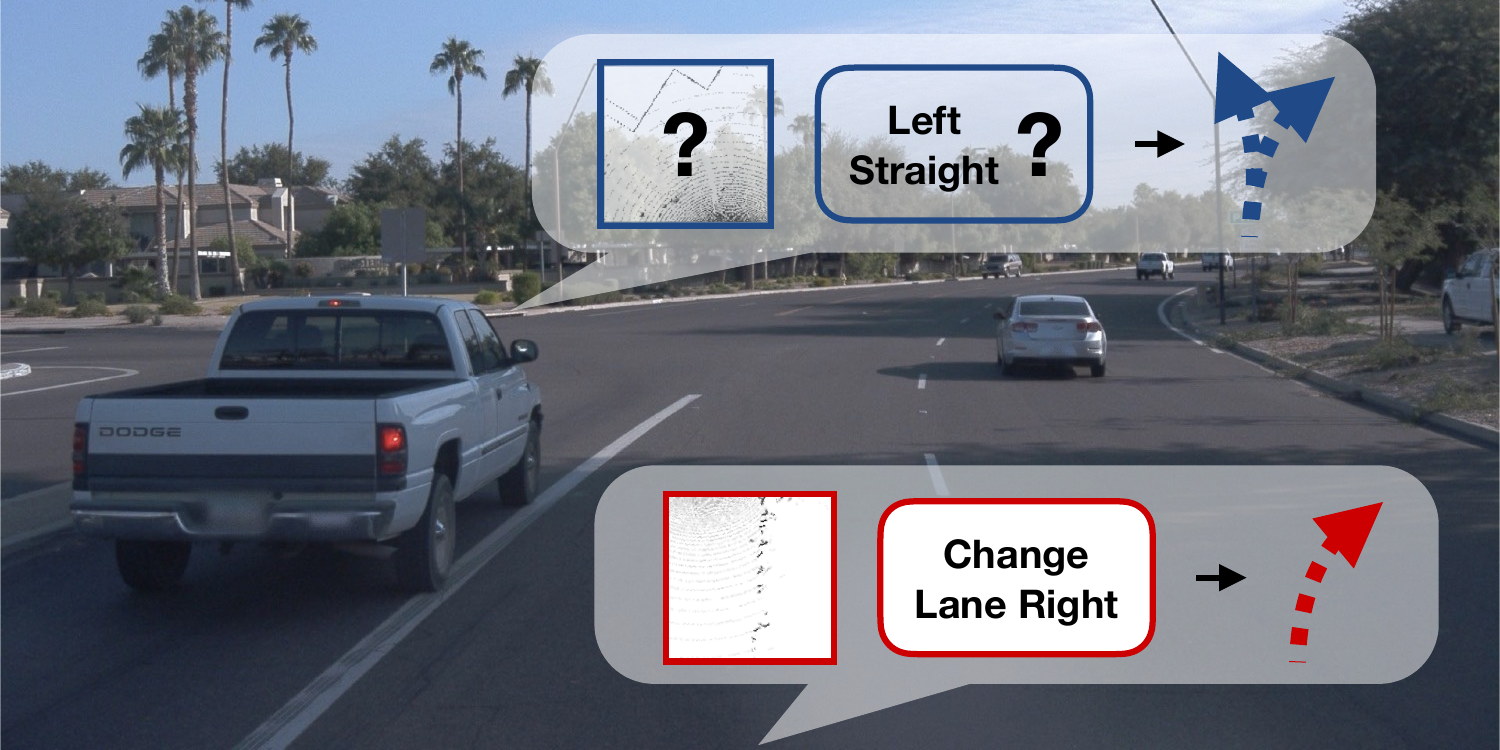}
\caption{We present LAV, a mapless, learning-based end-to-end driving system. LAV takes as input multi-modal sensor readings and learns from all nearby vehicles in the scene for both perception and planning. At test time, LAV predicts multi-modal future trajectories for all detected vehicles, including the ego-vehicle. Picture credit -- Waymo open dataset~\cite{sun2020scalability}.}
\vspace{-1.5em}
\end{figure}

We observe that, although many of us have not experienced traffic accidents ourselves, everyone has at least observed several accidents throughout our driving career. 
The same applies to safety-critical driving scenarios: While the data-collecting ego-vehicle might not experience accident-prone situations itself, it is likely its driving logs contain states that are interesting or safety-critical, but experienced by \textit{other vehicles}.
Training on other vehicles' trajectories helps not only with sample efficiency, but also greatly increase the chance that the model sees interesting scenarios.
Moreover, knowing other vehicles' future trajectories helps the ego-vehicle avoid collisions.

The main challenge with training on all vehicles lies in the partial observability of other vehicles.
Unlike the ego-vehicle, other vehicles have only partially observed motion trajectories, exposing no control commands or higher-level goals.
This makes direct training~\cite{codevilla2018end,chen2019lbc,chen2021learning,Prakash2021CVPR,Chitta2021ICCV} on other vehicles' traces close to impossible.
More importantly, other vehicles have no accessible sensors.
To learn from other vehicles, a model has to infer their surrounding state using the ego-vehicle's sensors.

Our framework, \textbf{L}earning from \textbf{A}ll \textbf{V}ehicles (LAV), handles the partial observability of both perception and motion in one joint recognition, prediction, and planning stack.
We decouple the partial observability challenge of perception and action using a privileged distillation approach~\cite{chen2019lbc}.
LAV first learns a perception model that outputs a viewpoint invariant representation using auxiliary supervision from 3D detection and segmentation tasks.
By definition, this auxiliary task does not distinguish between the ego-vehicle and other vehicles in the scene and thus learns a viewpoint invariant representation.
It handles the partial observability of sensors.
In parallel, LAV learns a privileged motion planner~\cite{chen2019lbc}.
Instead of predicting steering and acceleration, which are only available for the ego-vehicle, we use future waypoints to represent the motion plan.
We use ground-truth computer-vision labels as inputs to the privileged motion planner.
Computer-vision labels ensure viewpoint invariance, waypoints provide an invariant representation of motion.
The privileged motion planner predicts trajectories of all nearby vehicles and infers their high-level commands.
Finally, we combine the two models in a joint framework using privileged distillation~\cite{chen2019lbc}.
This final distillation learns a motion prediction model from all vehicles using the viewpoint invariant vision features of the perception model.
The distilled policy drives from raw sensor inputs alone.

We validate our method in the CARLA driving simulator~\cite{dosovitskiy2017carla}. 
At the time of submission, our method ranks first on the CARLA public leaderboard\footnote{\url{https://leaderboard.carla.org/leaderboard/}}.
It attains a $\mathbf{61.85}$ driving score and a $\mathbf{94.46}$ route completion rate.
Both are the highest among all methods and outperform the prior state-of-the-art method by a wide margin, increasing driving score and route completion rate by \textbf{25} and \textbf{24} points respectively.
Our method won the 2021 CARLA Autonomous Driving challenge\footnote{\url{https://ml4ad.github.io/}}.
Code available at \url{https://github.com/dotchen/LAV}.

\begin{figure*}[t]
\centering
\begin{subfigure}[b]{0.48\textwidth}
\raggedright
\includegraphics[width=0.95\textwidth,page=1]{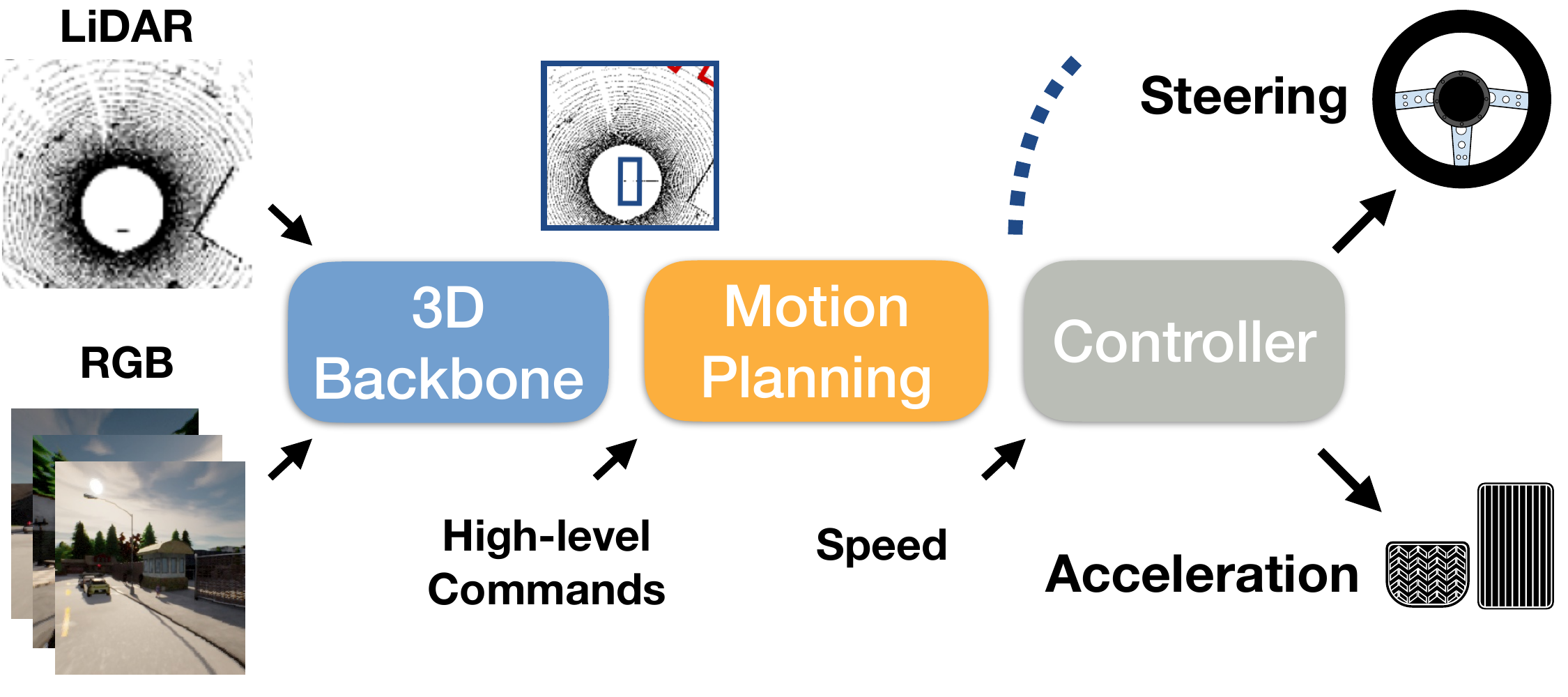}
\caption{Overview of the inference pipeline.}
\lblfig{overview_test}
\end{subfigure}%
\begin{subfigure}[b]{0.48\textwidth}
\raggedleft
\includegraphics[width=0.95\textwidth,page=2]{figures/pipeline_models.pdf}
\caption{Overview of the training pipeline for the motion planning module.}
\lblfig{overview_train}
\end{subfigure}
\caption{
Overview of the agent's pipeline. A 3D Backbone fuses LiDAR measurements and semantic segmentation from RGB cameras to produce a 2D spatial feature map.
This shared feature map serves as an input to a motion planner. 
At inference time (a), we use the central crop to predict the ego-vehicles trajectory.
At training time (b), we additionally use ground-truth detections of nearby vehicles to train a motion planner from all visible vehicles.
Detection results use rotated regions of interest (RoIs) of the shared feature map.
Finally, at inference time, a controller aggregates multiple motion predictions into a single steering and acceleration command.
}
\lblfig{architecture}
\vspace{-1em}
\end{figure*}

\section{Related Work}

\paragraph{Perception for autonomous driving} is driven by advances in visual understanding and recognition.
The perception system of a self-driving vehicle understands the scene by inferring its nearby objects and surrounding road structures.
A typical perception system takes as input LiDAR scans and performs object detection and tracking~\cite{zhou2020tracking,lang2019PointPillars,frossard2020strobe,zhou2020end,yin2021center}.
\citet{liang2018deep,vora2020pointpainting} fuse RGB camera and LiDAR scans for richer semantic information.
For roads, perception systems are categorized based on whether they require pre-recorded HD-Map: map-based systems localize themselves in the pre-recoded maps~\cite{levinson2007map,zheng2017high,ma2019exploiting};
mapless systems either perform online mapping~\cite{garnett20193d,guo2020gen,casas2021mp3}, or they implicitly predict road-related affordances~\cite{chen2015deepdriving,sauer2018conditional,toromanoff2020end}.
\citet{bansal2018chauffeurnet,zeng2019end} represent the perception outputs as bird's-eye-viewed (BEV) spatial grids;
more recently,~\citet{gao2020vectornet,li2021hdmapnet} represent perception outputs in a parameterized vector space for a more compact representation. 
Our approach takes multi-modal sensor data as input and performs online mapping and object detection.
However, we do not directly use the predicted map to perform classical planning.
Instead, we learn a planner from data using imitation learning.
This planner uses every vehicle it encounters on the road as a supervisory signal to enhance the diversity of the training data.

\paragraph{Behavior prediction} focuses on forecasting the future state of driving scenes.
In autonomous driving, a behavior predictor takes as either the input representations obtained from perception or raw sensor data; it predicts trajectories of the dynamic objects in the driving scene.
\citet{luo2018fast,zeng2019end} predict single, deterministic future trajectories of the detected vehicles.
\citet{zhao2020tnt,casas2018intentnet} model multi-modal future trajectories by using conditional models.
\citet{chai2019multipath} predicts trajectories as Gaussian mixtures to represent uncertainty in the euclidean space.
\citet{lee2017desire,cui2021lookout} use latent variables and VAEs to model actor and scene specific uncertainties.
Recently,~\citet{casas2021mp3,kamenev2021predictionnet,hu2021fiery} combine perception and behavior prediction by directly predicting the occupancy maps.
Our approach is highly related to the task of behavior prediction, as it also trains on all nearby vehicles' trajectories. 
Our approach consists of a behavior predictor. In particular, it applies a conditional motion planner on all nearby vehicles, including the ego-vehicle.

\paragraph{Learning-based motion planning} uses imitation learning or reinforcement learning to plan future trajectories.
Pioneered by~\citet{pomerleau1989alvinn}, imitation learning for autonomous driving regresses sensor inputs to controls by imitating the recorded expert trajectories.
\citet{codevilla2018end} use conditional branching and high-level commands to extend imitative models for urban driving.
\citet{zeng2019end} use imitation learning to train a cost volume predictor for planning;
\citet{chen2015deepdriving,sauer2018conditional} predict actions from the learned affordances.
\citet{chen2019lbc} uses on-policy distillation to handle distribution shift as well as to provide stronger imitative supervision signals.
Reinforcement learning, on the other hand, trains policies from a user-defined reward function.
\citet{kendall2019learning} train a lane following driving policy using DDPG;
\citet{toromanoff2020end} use distributed Rainbow-IQN to train an urban driving policy with competitive performance.
Recently,~\citet{chen2021learning} use model-based reinforcement learning and distillation to train a driving policy in an offline manner.
Our approach builds upon~\citet{chen2019lbc} and trains a motion planner using imitation learning and distillation.
However, unlike most prior methods, we train motion planning on data from all nearby vehicles in addition to the ego-vehicle.

Our idea of training the ego motion planner using data from all vehicles is closely related to~\citet{filos2021psiphi} and~\citet{zhang2021learning}.
\citet{filos2021psiphi} extends offline reinforcement learning to learn from other agents' behaviors.
\citet{zhang2021learning} train a privileged imitation learning policy that learns from other vehicles in a scene.
Their policy side-steps partial observability by training a policy that acts only on the ground truth state of the simulator.
It assumes perfect perception or access to other agents' sensors.
LAV, on the other hand, operates on raw sensor inputs and learns a viewpoint invariant intermediate representation.

\begin{figure*}[ht!]
\centering
\begin{subfigure}[b]{0.33\textwidth}
\raggedright
\includegraphics[width=0.95\textwidth,page=1]{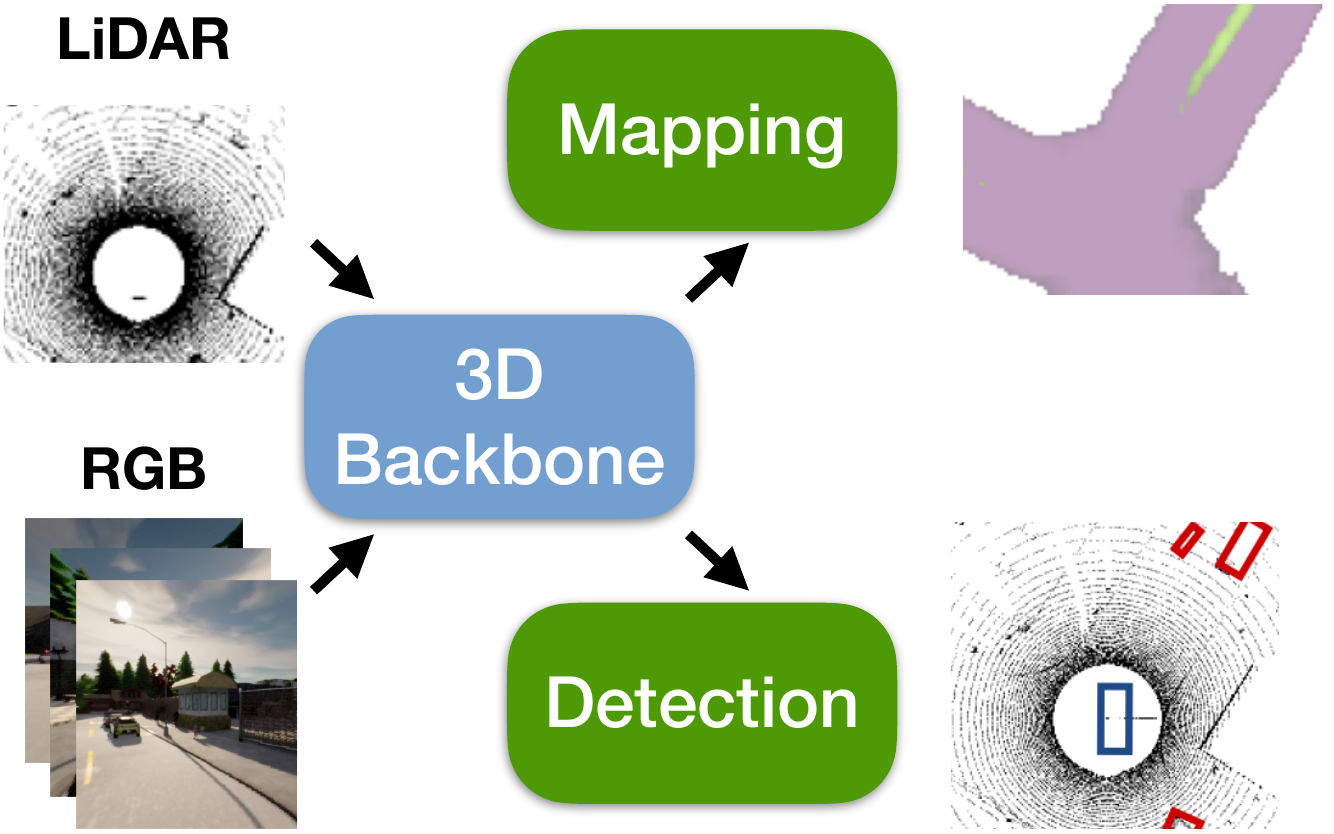}
\caption{Perception training.}
\lblfig{train_perception}
\end{subfigure}%
\begin{subfigure}[b]{0.33\textwidth}
\includegraphics[width=0.95\textwidth,page=2]{figures/pipeline_learning.pdf}
\caption{Privileged motion models training.}
\lblfig{train_motion}
\end{subfigure}
\begin{subfigure}[b]{0.33\textwidth}
\raggedleft
\includegraphics[width=0.95\textwidth,page=3]{figures/pipeline_learning.pdf}
\caption{Final distillation.}
\lblfig{train_distill}
\end{subfigure}
\caption{Overview of our training pipeline. 
\textbf{(a)}~We train a 3D perception model using detection and semantic mapping as the supervision signal. Both tasks help learn a viewpoint-invariant spatial representation. Detection additionally predicts other vehicles' poses which we use to forecast their future trajectories at inference.
The perception module produces a vehicle-independent feature representation used in motion planning.
\textbf{(b)}~In parallel, we train a motion planner over ground truth perception. We train the model using traces from all nearby vehicles using their future trajectory as supervision.
\textbf{(c)}~Finally, we combine the models learned in \textbf{(a)} and \textbf{(b)} using distillation. This model learns how all vehicles plan in an end-to-end manner using only the ego-vehicles sensor inputs.}
\label{fig:learning}
\vspace{-3mm}
\end{figure*}

\section{Learning from All Vehicles}

We aim to build a deterministic driving model $\pi$ that at each timestep $t$ maps sensor readings, high-level navigational command, and vehicle state to raw control command $a_t$.
We opt for an end-to-end differentiable three-stage modular pipeline: A perception module, a motion planner, and a low-level controller.
See \reffig{overview_test} for an overview.

The perception module is trained from massive labeled supervision with two goals in mind: To create a robust and generalizable representation of the surrounding world, and to build vehicle-invariant features that help supervise the motion planner.
\refsec{backbone} describes the overall architecture and training setup of the perception module.
It maps raw sensor readings to a map-view feature representation.

The motion planner uses the map-view features of the perception model to produce a series of waypoints describing the future trajectory of the vehicles.
Motion planners commonly use supervision from just the ego-vehicle for this prediction~\cite{chen2019lbc}.
This supervision is quite sparse and provides the motion planner with just a single series of labels per collected data point.
In our framework, we learn motion planning from all vehicles that surround the ego-vehicle.
This is possible because our perception system produces vehicle-invariant features as inputs; it is also because the outputs of the motion planner, the future trajectories, can be easily obtained from ground truth driving logs.
\reffig{overview_train} shows an overview of the motion planner training.
\refsec{mplan} describes the motion planner and its training setup.

Finally, a low-level controller converts motion plans into actual steering and acceleration commands that are executed on the vehicle.
At test time, the low-level controller considers other vehicles' motion plans to make emergency stop decisions.
\refsec{control} describes the controller.

\subsection{A vehicle-independent perception model}
\lblsec{backbone}

The core objective of any perception module is to build an intermediate representation that readily generalizes from training to test conditions.
In our setup, a secondary goal is to build input features to the motion planner that are indistinguishable between the current vehicle and nearby vehicles.
The closer the output representations of the ego-vehicle and other vehicles are, the better motion plans transfer between those vehicles.
Here, we opt for a metric map-based output representation.
In a metric map, rotated ROI pooling extracts fixed-sized feature representations for training vehicles.

Specifically, we use three RGB cameras $\mathbf{I}_t = \{I_t^1, I_t^2, I_t^3\}$ surrounding the vehicle and one LiDAR sensor $L_t$ as an input.
We combine the color and LiDAR inputs using point-painting~\cite{vora2020pointpainting} from RGB inputs and a light-weight CenterPoint~\cite{yin2021center} with PointPillars~\cite{lang2019PointPillars} 3D backbone.
The backbone provides us with a map-view feature representation $f \in \mathbb{R}^{W \times H \times C}$ of width $W$ and height $H$ with $C$ channels.

We train the backbone network using a combination of semantic segmentation and detection losses.
See \reffig{train_perception} for an overview.
For every pixel in map-view, we predict a road mask, solid and broken lane boundaries.
We use a binary classifier, and binary cross-entropy loss, as road and lane-marking can overlap.
In addition, we train a CenterPoint-style detector~\cite{yin2021center} for pedestrians and vehicles.
Most importantly, we explicitly label the ego-vehicle in this detector.
This minimizes the feature distance between ego-vehicle and other vehicles and enables better transfer.
We pre-train the perception model using fully labeled data and use rotation augmentations around the ego-vehicle to increase the robustness of the learned model.

Supervised pre-training has two advantages.
It generalizes better to unseen test conditions.
It also learns a similar feature representation for all vehicles.
This feature representation is next used in the motion planner.

\subsection{Learning to plan motion from all vehicles}
\lblsec{mplan}
The motion planner uses the output of the perception system to predict a series of future waypoints describing positions the vehicles should steer towards.
Here, we propose a novel two-stage motion planner that combines geometric GPS targets and discrete high-level commands.
We use a standard RNN formulation~\cite{lee2017desire,Prakash2021CVPR} to predict $n=10$ future waypoints $y_1, \ldots, y_n \in \mathbb{R}^2$. We use $n=20$ for our second leaderboard submission.
The motion planner uses a high-level command $c$ and intermediate GNSS coordinate goal $g \in \mathbb{R}^2$ to perform different driving maneuvers.
In CARLA, GNSS goals are sampled every 50-100 meters and contain a measurement error of around one meter.
Possible high-level commands $c$ include \textit{turn-left}, \textit{turn-right}, \textit{go-straight}, \textit{follow-lane}, \textit{change-lane-to-left}, \textit{change-lane-to-right}.

Let $M(\hat f, c): \rightarrow \mathbb{R}^{n \times 2}$ be the motion planner conditioned on high-level command $c$ and warped features $\hat{f}$ for the Region of Interest (RoI) at the location and orientation of the vehicle in question.
For all vehicles, we observe their future trajectory to obtain supervision for future waypoints $y$.
For the ego-vehicle, the simulator provides a ground truth high-level command $\hat c$ and provides sufficient supervision to train the motion planner
\begin{align}
\mathcal{L}_M^{ego} = \mathrm{E}_{\hat f, y, \hat c}\left[\|y - M(\hat f, \hat  c)\|_1\right].\lbleq{m_ego}
\end{align}
However, other vehicles do not expose their high-level commands.
While it may be possible to infer this command from future trajectories alone, any rule-based inference will be ambiguous and noisy.
We instead allow the model to infer the high-level command directly and optimize the plan for the most fitting high-level command.
\begin{align}
\mathcal{L}_M^{other} = \mathrm{E}_{\hat f, y}\left[\min_c\|y - M(\hat f, c)\|_1\right].\lbleq{m_other}
\end{align}
At training time we optimize both losses $\mathcal{L}_M^{ego}$ and $\mathcal{L}_M^{other}$ jointly.
To avoid mode collapse, we use an adaptive weight $\lambda$ on $\mathcal{L}_M^{other}$ which starts small and converges to 1 as training progresses.
We found $\lambda = 1 - 0.8^{it/4000}$ to work well in practice.
This is only applied during the privileged training stage.

The resulting motion planner $M$ finds good coarse trajectories for a wide range of traffic scenarios.
It learns to plan for all vehicles it sees.
However, the resulting motion plan may be noisy as high-level commands $c$ are ambiguous.

In a second stage, we refine the motion plan using an additional RNN-based motion planning network $M^\prime(\hat f, g, \tilde y) \in \mathbb{R}^{n \times 2}$.
The motion refinement network uses the same ROI-warped feature $\hat f$, the previously predicted motion plan $\tilde y$, and the more fine-grain GNSS goal $g$ as input.
We normalize $g$ in the ego-vehicle's coordinate.
It then produces a delta to the original trajectory as output.
Since GNSS goals are only available for the ego-vehicle, we train the refinement $\hat M$ only on ego-vehicle trajectories
\begin{align}
\mathcal{L}_M^{refine} = \mathrm{E}_{\hat f, y, \tilde y, \hat g}\left[\|\tilde y + M^\prime(\hat f, \hat g, \tilde y) - y\|_1\right].\lbleq{m_refine}
\end{align}
To increase robustness of the refinement model, we use the output of $M$ from all high-level commands during training. 
During both training and testing, we roll out the same refinement network multiple times to recursively refine the predicted trajectory.
We use the refined trajectories for planning unless the high-level command is lane-changing.
The above loss then applies to each step of the rollout.

In practice, we learn the motion planner in a privileged distillation framework~\cite{chen2019lbc}.
See \reffig{train_motion} and \reffig{train_distill} for an overview.
We first learn motion planning on ground truth trajectories and ground-truth perception outputs and regions of interest using the losses \eqref{eq:m_ego}-\eqref{eq:m_refine}.
We then use the privileged motion planner to supervise a motion planner that uses the inferred perception outputs.
During this second stage, we supervise predictions on all high-level commands which leads to a richer supervisory signal~\cite{chen2019lbc}.
We additionally distill a high-level command classifier for other vehicles which we use later in the vehicle-aware controller.
This stage trains end-to-end by backpropagating gradients from motion prediction and planning to the perception backbone, allowing perception models to attend to the low-level details in the scenes.
We keep the pre-training perception loss in the previous stage as an auxilliary supervision to regularize the features.

\subsection{Vehicle-aware control}
\lblsec{control}

The controller translates a motion plan into actual driving commands.
We use two PID controllers for latitudinal (steering) and longitudinal (acceleration) control.
Both PID controllers use basic statistics of the refined motion plan as an input to produce a continuous output command.
The longitudinal PID controller additionally uses the current speed as an input to compute acceleration.
We overwrite braking using a separate neural network classifier $B$ in case of traffic light and hazard stoppages.
The classifier uses the same image inputs as the perception module plus one additional camera with telephoto lenses to capture distant traffic lights.
The classifier learns the braking behavior of the data-collecting ego-vehicle using recorded brake actions.
Finally, we reuse the motion plans learned from other vehicles to detect potential collisions and perform hazard stops.
Specifically, we use the 3D detections of the backbone to find all nearby vehicles.
For each, we use the motion planner $M$ to produce future trajectories over each high-level command.
We use all motion plans above the high-level command likelihood threshold to check for collisions with the ego-vehicle's motion plan.

\section{Implementation details}
\paragraph{Perception.}
We use PointPillars~\cite{lang2019PointPillars} with PointPainting~\cite{vora2020pointpainting} as our multi-modal 3D perception backbone $P_B$.
In particular, given RGB images captured from three frontal facing camera $\{I^0_t,I^1_t,I^2_t\}$ with extrinsic matrices $E=\{E^0,E^1,E^2\}$, we use a ERFNet~\cite{romera2017erfnet} to compute their semantic segmentation scores $\mathbf{S}_t =\{S^0_t,S^1_t,S^2_t\}$.
We use five semantic classes: ``background'', ``vehicles'', ``roads'', ``lane markings'' and ``pedestrians''.
For each LiDAR point $l \in L_t$, we use PointPainting~\cite{vora2020pointpainting} to concatenate its corresponding semantic classes using the segmentation scores:
$l^s_t = \text{PointPaint}(\mathbf{S}_t, \mathcal{T}_{E},l_t)$.
$\mathcal{T}_{E}$ is the perspective transform function.

For PointPillars, we use FC-64-64 with BatchNorm~\cite{ioffe2015batch} as its PointNet.
We create pillars for LiDAR points for $x \in [-10\text{m}, 70\text{m}]$ and $y \in [-40\text{m}, 40\text{m}]$.
Each pillar represents a $0.25\text{m} \times 0.25\text{m}$ spatial region.
We use the default 2D CNNs with multi-scale features to obtain the spatial features $\phi_t \in \mathbb{R}^{192\times160 \times 160}$ with 
$0.5\times$ resolution of the original pillars. 
Unlike the original PointPillars which directly builds dense pillars specified by the hyperparameters, we sparsely represent the pillars.
We also use a sparse PointNet to process the corresponding sparse pillar features.
This allows us to process all pillars efficiently both in space and time.

We use a branching architecture for the detection and mapping heads.
We use a simplified one-stage CenterPoint~\cite{zhou2019objects, yin2021center} formulation for BEV object detection.
In particular, we predict two ``centerness'' maps, one for vehicles and one for pedestrians; we also predict an orientation and bounding-box maps that are both class-agnostic.
For mapping, we predict a BEV semantic map for roads, solid lane markings and broken lane markings.
Each map is generated using a separate $3\times3$ convolution followed by a $3\times 3$ up-convolution with stride $2$, all from the shared backbone $P_B$.
At test time, we use a 2D max pooling layer as a simplified version of NMS.

We additionally train a binary brake classifier that takes as input all the four camera RGB images.
We feed the telephoto lens image and the stitched other three images to a ResNet-18 followed by a global average pooling layer. This gives us fixed-sized embeddings of $z_1, z_2 \in \mathbb{R}^{512}$. We concatenate $z_1, z_2$ and feed it to a linear layer to predict the binary brake.

\paragraph{Prediction and Planning.}
Given the ego-vehicle and a list of vehicle detection, we use differentiable warping to crop a rotated region of interest (RoI) $\hat f^i$ for each vehicle location and yaw angle.
A CNN followed by global average pooling takes as input the rotated RoI features and returns a fixed-sized embedding $z^i$ for each vehicle $i$.
$z^i$ is shared among $M$ and $M^\prime$.
The motion planner $M$ uses a separate GRU~\cite{cho2014learning} for each high-level command.
The GRU is rolled out $n$ times to produce an offset between consecutive waypoints.
The refinement motion planner $M^\prime$ uses two forms of recursions and rollouts: rollouts along waypoint and rollouts along refinement iterations.
It predicts an offset from the prior motion plan for each iteration.
The refinement motion plan relies on just a single GRU unit that takes the GNSS goal $g$ as an additional input.
Both motion planners use a linear layer to transform GRU states into the desired outputs.

\begin{figure}[t]
\centering
\includegraphics[width=0.95\linewidth]{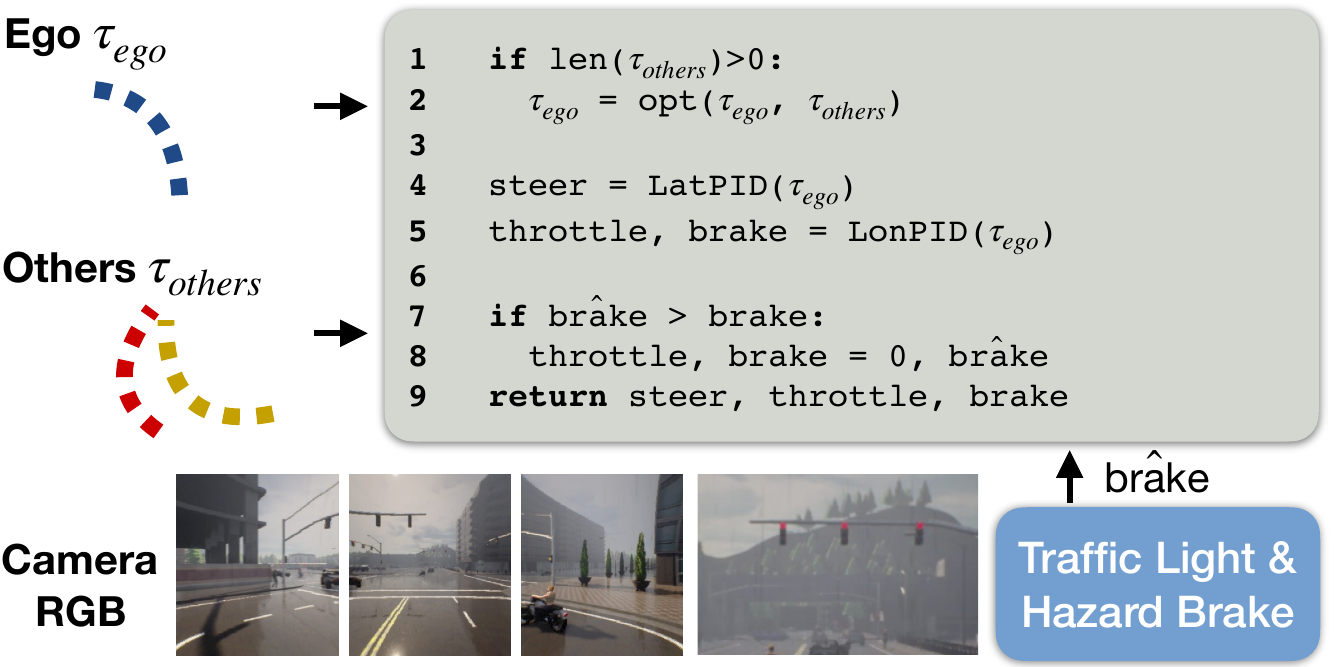}
\caption{Overview of our controller logic. The controller considers all vehicles and their predicted multi-modal future trajectories. It additionally uses an image-only brake predictor to handle traffic sign and hazard stoppages.}
\vspace{-1.5em}
\lblfig{controller}
\end{figure}

\paragraph{Control.}
The controller $C$ takes as input refined ego-trajectory $\tau = M^\prime(\hat f, \tilde y, \hat g)$.
See~\reffig{controller} for an overview.
If predicted trajectory $\tau$ leads to a collision with other traffic participants, we adjust it.
For now we perform a hard stop using a hard-coded braking logic.
If the predicted trajectory is collision free, we follow it directly.
We use two PID controllers for latitudinal and longitudinal control respectively.
For latitudinal control, we use the $5$-th point in $\tau_5$ as the aim point to compute the steering error.
For longitudinal control, we use the difference between target speed inferred from $\|\tau_{k+1}-\tau_k\|$ and the current speed $v_t$ to compute acceleration.
We use $K_P=1.0, K_I=0.5, K_D=0.2$ for the latitudinal PID controller, and we use $K_P=5.0,K_I=0.5,K_D=1.0$ for the longitudinal PID controller.
We overwrite the brake control with the predicted brake score if it is larger than the brake computed from the longitudinal controller.

\begin{table}[t]
\centering
\begin{tabular}{l@{\ }c@{\ \ }c@{\ \ }c@{\ \ }c@{\ \ }}
\toprule
    Rank & Method & \thead{Driving \\ Score } & \thead{Route \\ Completion} & \thead{Infraction \\ Score} \\
    \cmidrule(r){1-2}
    \cmidrule(r){3-5}
    $1$ & \textbf{LAV} & $\textbf{61.85}$ & $\textbf{94.46}$ & $0.64$  \\
    $2$ & GRIAD~\cite{chekroun2021gri} & $36.79$ & $61.85$ & $0.60$ \\
    $3$ & TransFuser+~\cite{jaeger2021master} & $34.58$ & $69.84$ & $0.56$ \\
    $4$ & Rails~\cite{chen2021learning} & $31.37$ & $57.65$ & $0.56$ \\
    $5$ & IARL~\cite{toromanoff2020end} & $24.98$ & $46.97$ & $0.52$ \\
    $6$ & NEAT~\cite{Chitta2021ICCV} & $21.83$ & $41.71$ & $0.65$ \\
    $7$ & TransFuser~\cite{Prakash2021CVPR} & $16.93$ & $51.82$ & $0.42$ \\
    $8$ & LBC~\cite{chen2019lbc} & $8.94$ & $17.54$ & $\textbf{0.73}$ \\
    \bottomrule
\end{tabular}
\caption{Comparison of the driving score (main metric), route completion and infraction score on the public CARLA leaderboard~\cite{leaderboard} (accessed Jan 2022). All three metrics are higher the better. We outperform prior methods by a wide margin. Detailed infraction numbers reported in the supplement for reference.
}
\vspace{-1em}
\lbltbl{sota}
\end{table}

\begin{table*}[t]
\centering
\begin{tabular}{l@{\ \ \ \ \ }c@{\ \ \ \ }c@{\ \ \ \ }c@{\ \ \ \ }c@{\ \ }c@{\ \ }c@{\ \ }c@{\ \ }}
    \toprule
     & \thead{Driving \\ Score } & \thead{Route \\ Completion} & \thead{Infraction \\ Score} & \thead{Vehicle \\ Collisions} & \thead{Pedestrian \\ Collisions} & \thead{Layout \\ Collisions} & \thead{Red light \\ Violations} \\
    \cmidrule(r){1-1}
    \cmidrule(r){2-8}
    \textbf{LAV} & $\textbf{45.20} \pm 6.35$ & $\textbf{91.55} \pm 5.61$ & $0.49 \pm 0.06$ & $\textbf{0.92} \pm 0.42$ & $0.00 \pm 0.00$ & $\textbf{0.33} \pm 0.50$ & $0.28 \pm 0.28$ \\
    \cmidrule(r){1-1}
    \cmidrule(r){2-8}
    Ego-vehicle only & $38.56 \pm 1.86$ & $84.76 \pm 5.12$ & $0.46 \pm 0.02$ & $1.17 \pm 0.50$ & $0.00 \pm 0.00$ & $1.82 \pm 0.06$ & $0.34 \pm 0.20$ \\
    No distillation & $28.23 \pm 2.27$ & $81.05 \pm 6.04$ & $0.36 \pm 0.04$ & $2.08 \pm 0.34$ & $0.00 \pm 0.00$ & $7.87 \pm 0.15$ & $\textbf{0.21} \pm 0.04$ \\
    \bottomrule
\end{tabular}
\caption{Driving performance ablation of the key components of our approach on test towns. Infractions are measured as number of occurrences per kilometer traveled. Mean and standard deviation are computed over three runs.}
\vspace{-1em}
\lbltbl{key_ablate}
\end{table*}

\section{Experiments}

We evaluate our method on the CARLA simulator~\cite{dosovitskiy2017carla} using closed-loop driving.
We compare our approach with the state-of-the-art methods on the public leaderboard, and we perform ablation study on the effect of our design choices locally.
For our online leaderboard submission, we train on all the 8 publicly available towns using a dataset of 400K frames, collected with the CARLA behavior agent under randomized weathers.
For ablations, we only train on 4 out of the 8 towns, resulting in a dataset of 186K frames.
We test on two other unseen towns.

More details of the dataset statistics are provided in the supplement for reference.

\subsection{Comparison with state-of-the-art}
\reftbl{sota} compares our method with prior state-of-the-art methods on the CARLA public leaderboard~\cite{leaderboard}.
The CARLA leaderboard evaluates autonomous driving systems under unseen and partially adversarial conditions.
Vehicles are tasked to complete a set of predefined routes in new towns.
For each route, the simulator adds dangerous scenarios such as suddenly crossing pedestrians or aggressive lane-changing vehicles.
These scenarios are modeled after the NHTSA typology~\cite{leaderboard}.
The leaderboard measures how far self-driving vehicles proceed along a route within a fixed time budget, and how often they cause traffic infractions.
In~\reftbl{sota}, we list three key metrics of the leaderboard: Driving Score, Route Completion, and Infraction Score.
Route Completion measures the distance percentage an agent is able to complete;
Infraction Score measures how often an agent drives without causing infractions;
Driving Score measures route completion rate weighted by infractions per route.
Driving Score and Route Completion are the two main metrics of comparison.
A vehicle standing perfectly still will receive an infraction score of $1$.
All three metrics are higher the better.
We refer readers to the official leaderboard~\cite{leaderboard} for a more detailed description of the metrics.

We compare to the top 7 entries on the leaderboard.
GRIAD~\cite{chekroun2021gri} and Transfuser+ are unpublished, concurrent submissions.
Rails~\cite{chen2021learning} is a model-based reinforcement-learning method that trains vision-based driving policies from offline driving logs.
IARL~\cite{toromanoff2020end} is based on state-of-the-art model-free reinforcement-learning with distributed training.
NEAT~\cite{Chitta2021ICCV} uses imitation-learning with attention and implicit functions.
Transfuser~\cite{Prakash2021CVPR} uses imitation-learning with attention-based sensor fusion.
LBC~\cite{chen2019lbc} relies on knowledge distillation with imitation learning.
LBC is the closest comparison to our approach, as we also use knowledge distillation and imitation learning as a supervisory signal.
However, LAV additionally uses other observed vehicles to train the control policy.

LAV ranks first on the leaderboard, and it outperforms the prior leading entry by a wide margin.
It achieves $\mathbf{61.85}$ driving score, the highest among all methods, and $\mathbf{25}$ points higher than on the previous state-of-the-art GRIAD.
It also achieves a $\mathbf{94.46}$ Route Completion, the highest among all methods and $\mathbf{32}$ points higher than the previous state-of-the-art.
Moreover, previous top methods, such as Rails and IARL, require 1M and 40M frames to train the policies.
Our method uses only 400K training frames.
Our approach has a relatively high infraction score; however, we note that higher Route Completion naturally leads to more infractions.
A vehicle that drives slowly or stands still causes fewer infractions but struggles to complete its routes.
See LBC for example.

\subsection{Ablation study}

We answer few important questions on our design choices.
We again evaluate on Driving Score, Route Completion and Infraction Score.
However, we cannot use the online leaderboard directly for these additional experiments.
We instead use a local setup with similar characteristics to the Leaderboard.
In particular, we train on 4 out of 8 towns (Town01, Town03, Town04 and Town06), and evaluate on 2 unseen towns (Town02 and Town05).
We select 4 representative routes, 2 from each town, and we evaluate each route with 4 weathers: ``Clear Noon'', ``Cloudy Sunset'', ``Soft Rain Dawn'' and ``Hard Rain Night''.
We evaluate each setup for 3 runs and report the mean and standard deviation.
This results in 48 trials for each model.
All ablated models use a similar but slightly outdated setup as our leaderboard entry.
The main differences are: 1. the ablated models use U-Net as the semantic segmentation backbone 2. the ablated models use FC-32-32 in PointPillars and 3. slightly different controller hyperparameters.

\reftbl{key_ablate} studies the effects of our key design choices.
We compare to two variants of our approach: 1) One that only trains on ego-vehicle data, 2) One that does not perform privileged distillation.
We find that training on other vehicles' trajectories and viewpoints results in lower performance on both route completion and infraction.
The performance degradation is smaller than expected likely because our auxiliary perception supervision makes the motion models generalize well to distribution shifts, caused by both test time errors and viewpoints changes.
Not using privileged distillation results in a larger performance drop.
Without distillation, the motion models need to train from both noisy inputs and labels, thus tackling a much harder learning problem. 
Our full model achieves the highest scores in all three metrics.

\reftbl{radius_ablate} studies the degree to which training on other vehicles' experiences affect the driving performance.
We evaluate three standards, where we only train vehicles within 5, 15 and 25 meters within the ego-vehicle, with 15 meters being our default value.
$\leq 5$m and $\leq 15$m performs equally well, while $\leq 25$m is slightly worse.
We think this is due to the fact that vehicles at range have too different of an appearance in the sensor inputs which the auxiliary supervision is unable to correct for.
The LiDAR sensor in CARLA mimics the Velodyne-64 rays model, which produces at most a few dozen measurements for cars at a distance of $25$m.

\reftbl{staged_ablate} studies different perception training schemes.
We compare the default staged training scheme (staged) with two variants of joint training: 1) No perception pre-training (No Pretrain.), and joint perception and motion training (Joint).
The first variant only optimizes the distillation loss, and the latter optimize perception and distillation simultaneously.
Both variants do not freeze the 3D backbone.
As expected, models without perception training perform poorly.
They suffer from the distribution shifts caused by viewpoint changes.
Joint training also performs worse than staged straining, because solving perception and planning simultaneously is harder than solving them in a disentangled manner, as also observed by~\citet{chen2019lbc}.

\reftbl{refinement_ablate} studies the effect of our iterative refinement module.
$K=0$ means we directly use the trajectories predicted by the motion planner $M_f$ to drive.
The default option $K=5$ performs the best, showing the benefits of iterative motion refinement.
Iterative refinement allows the model to elastically figure out what residuals to learn.
It also naturally combines the semantic information from the high-level command and the geometric information from the goals.

Detailed infraction numbers for~\reftbl{radius_ablate},~\reftbl{staged_ablate} and~\reftbl{refinement_ablate} are provided in the supplement for reference.

\begin{table}[t]
\centering
\begin{tabular}{l@{\ \ }c@{\ \ \ \ }c@{\ \ \ \ }c@{\ \ \ \ }}
    \toprule
    \thead{Vehicles \\ Range} & \thead{Driving \\ Score } & \thead{Route \\ Completion} & \thead{Infraction \\ Score} \\
    \cmidrule(r){1-1}
    \cmidrule(r){2-4}
    $\leq$ 5m  & $\textbf{46.06} \pm 1.70$ & $88.77 \pm 1.01$ & $0.51 \pm 0.02$ \\
    $\leq$ 15m & $45.20 \pm 6.35$ & $\textbf{91.55} \pm 5.61$ & $0.49 \pm 0.06$ \\
    $\leq$ 25m & $37.42 \pm 3.09$ & $89.56 \pm 5.61$ & $\textbf{0.61} \pm 0.12$ \\
    \bottomrule
\end{tabular}
\vspace{-0.5em}
\caption{Driving performance in test towns of models trained with different range of other vehicles. All models are the same except for other vehicles' maximum range used during training.}
\vspace{-0.5em}
\lbltbl{radius_ablate}
\end{table}

\begin{table}[t]
\centering
\begin{tabular}{c@{\ \ }c@{\ \ \ \ }c@{\ \ \ \ }c@{\ \ \ \ }}
    \toprule
    \thead{Perception \\ Training} & \thead{Driving \\ Score } & \thead{Route \\ Completion} & \thead{Infraction \\ Score} \\
    \cmidrule(r){1-1}
    \cmidrule(r){2-4}
    No Pretrain.   & $8.47  \pm 0.83$ & $9.34  \pm 0.35$ & $\textbf{0.90} \pm 0.07$ \\
    Joint  & $28.36 \pm 2.11$ & $79.58 \pm 4.99$ & $0.34 \pm 0.02$ \\
    Staged & $\textbf{45.20} \pm 6.35$ & $\textbf{91.55} \pm 5.61$ & $0.49 \pm 0.06$ \\
    \bottomrule
\end{tabular}
\vspace{-0.5em}
\caption{Driving performance in test towns of models with different perception training scheme. All models are the same except for perception training.}
\vspace{-0.5em}
\lbltbl{staged_ablate}
\end{table}

\begin{table}[t]
\centering
\begin{tabular}{c@{\ \ }c@{\ \ \ \ }c@{\ \ \ \ }c@{\ \ \ \ }}
    \toprule
    \thead{Refinement \\ Iteration} & \thead{Driving \\ Score } & \thead{Route \\ Completion} & \thead{Infraction \\ Score} \\
    \cmidrule(r){1-1}
    \cmidrule(r){2-4}
    $K=0$ & $12.69 \pm 2.86$ & $35.85 \pm 2.91$ & $0.42 \pm 0.03$ \\
    $K=1$ & $21.30 \pm 1.10$ & $85.90 \pm 2.46$ & $0.25 \pm 0.01$ \\
    $K=5$ & $\textbf{45.20} \pm 6.35$ & $\textbf{91.55} \pm 5.61$ & $\textbf{0.49} \pm 0.06$ \\
    \bottomrule
\end{tabular}
\vspace{-0.5em}
\caption{Driving performance ablation on the effect of motion refinement. All models are the same except for number of refinement iterations.}
\vspace{-1.5em}
\lbltbl{refinement_ablate}
\end{table}

\begin{figure*}[t]
\centering
\begin{subfigure}[b]{0.24\linewidth}
 \centering
 \includegraphics[width=\textwidth]{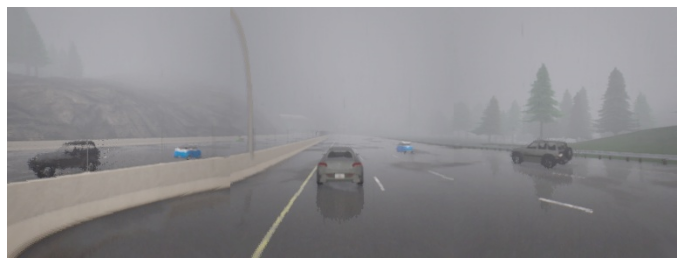}\vspace{1ex}
 \includegraphics[width=\textwidth]{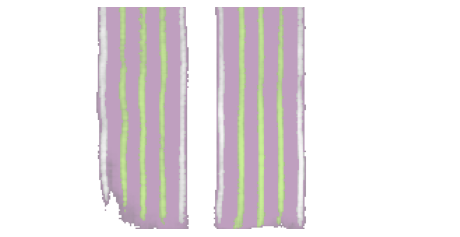}\vspace{1ex}
 \includegraphics[width=\textwidth]{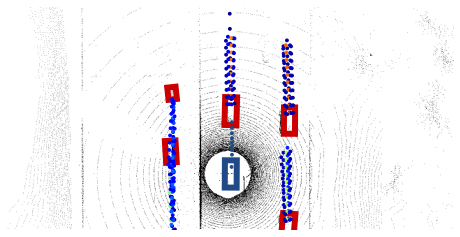}\vspace{1ex}
\end{subfigure}
\begin{subfigure}[b]{0.24\linewidth}
 \centering
 \includegraphics[width=\textwidth]{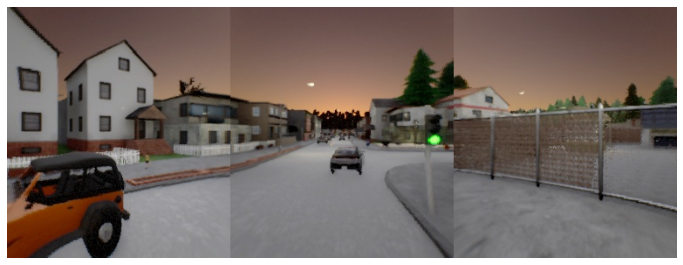}\vspace{1ex}
 \includegraphics[width=\textwidth]{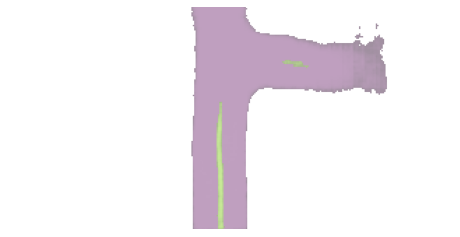}\vspace{1ex}
 \includegraphics[width=\textwidth]{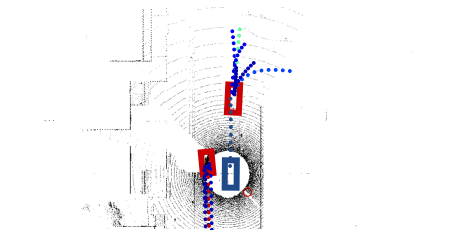}\vspace{1ex}
\end{subfigure}
\begin{subfigure}[b]{0.24\linewidth}
 \centering
 \includegraphics[width=\textwidth]{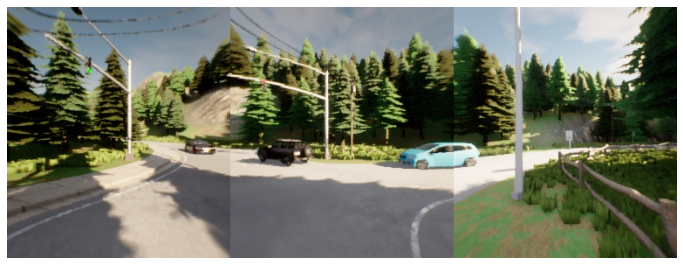}\vspace{1ex}
 \includegraphics[width=\textwidth]{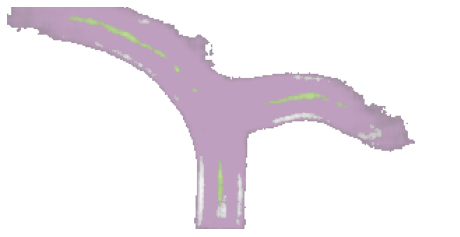}\vspace{1ex}
 \includegraphics[width=\textwidth]{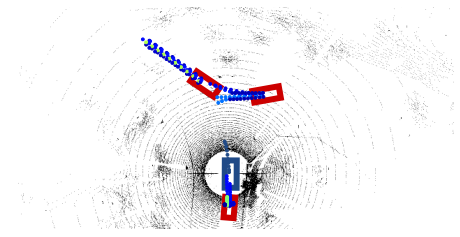}\vspace{1ex}
\end{subfigure}
\begin{subfigure}[b]{0.24\linewidth}
 \centering
 \includegraphics[width=\textwidth]{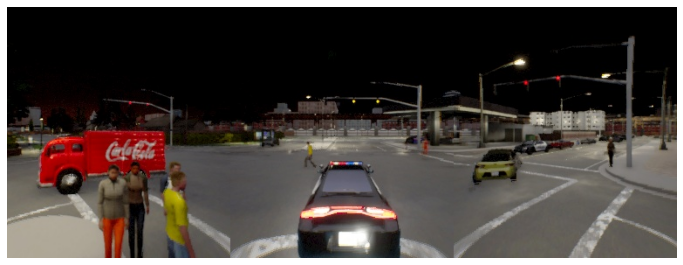}\vspace{1ex}
 \includegraphics[width=\textwidth]{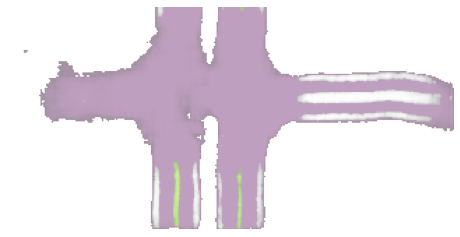}\vspace{1ex}
 \includegraphics[width=\textwidth]{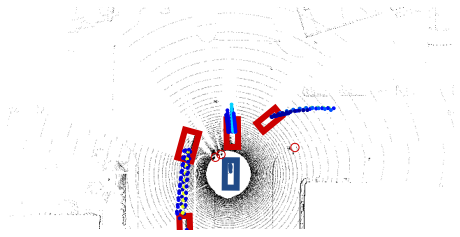}\vspace{1ex}
\end{subfigure}
\begin{subfigure}[b]{0.24\linewidth}
 \centering
 \includegraphics[width=\textwidth]{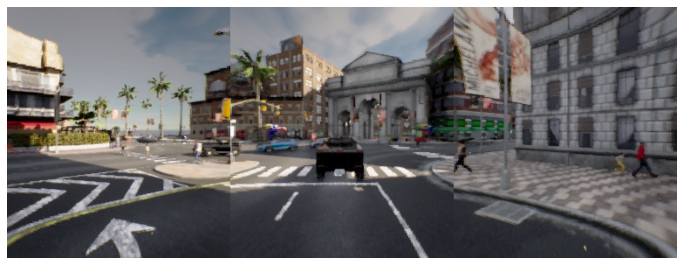}\vspace{1ex}
 \includegraphics[width=\textwidth]{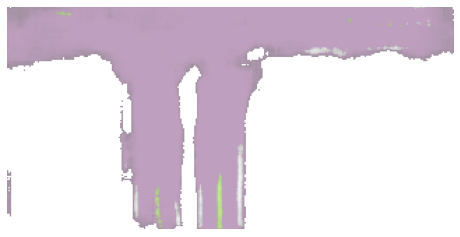}\vspace{1ex}
 \includegraphics[width=\textwidth]{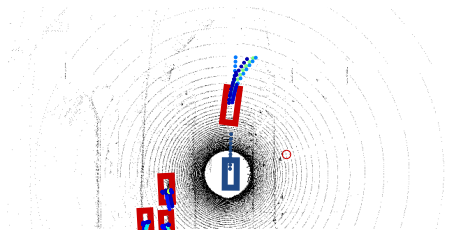}\vspace{1ex}
\end{subfigure}
\begin{subfigure}[b]{0.24\linewidth}
 \centering
 \includegraphics[width=\textwidth]{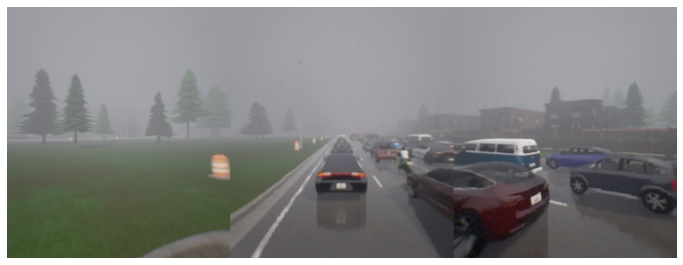}\vspace{1ex}
 \includegraphics[width=\textwidth]{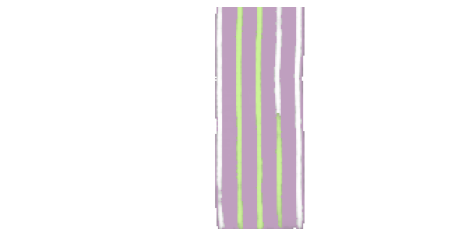}\vspace{1ex}
 \includegraphics[width=\textwidth]{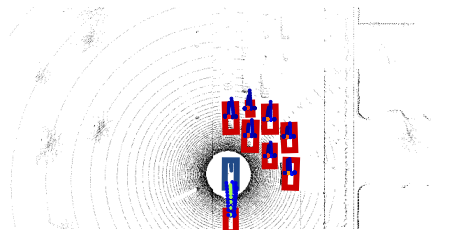}\vspace{1ex}
\end{subfigure}
\begin{subfigure}[b]{0.24\linewidth}
 \centering
 \includegraphics[width=\textwidth]{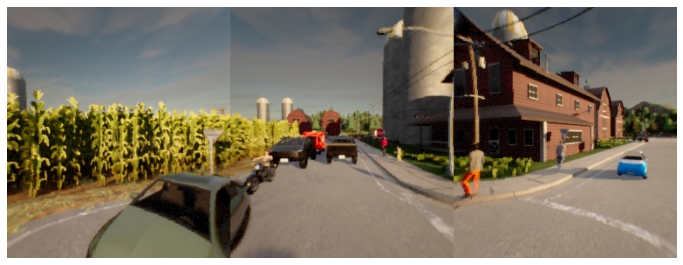}\vspace{1ex}
 \includegraphics[width=\textwidth]{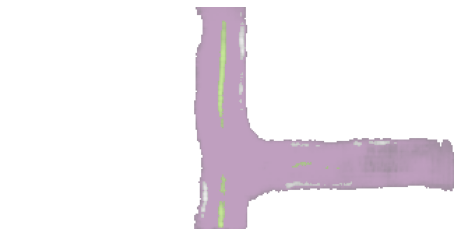}\vspace{1ex}
 \includegraphics[width=\textwidth]{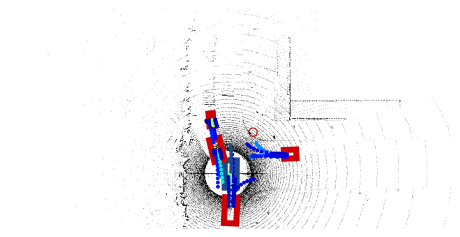}\vspace{1ex}
\end{subfigure}
\begin{subfigure}[b]{0.24\linewidth}
 \centering
 \includegraphics[width=\textwidth]{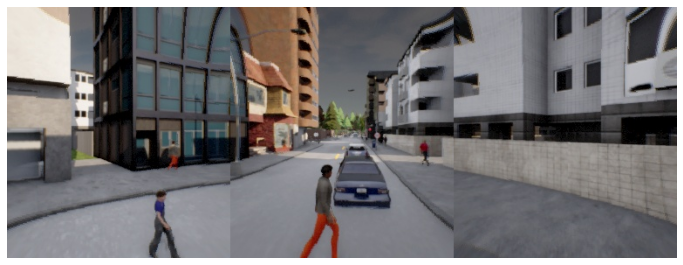}\vspace{1ex}
 \includegraphics[width=\textwidth]{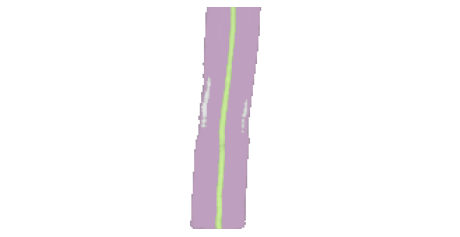}\vspace{1ex}
 \includegraphics[width=\textwidth]{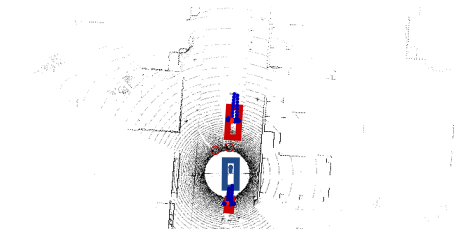}\vspace{1ex}
\end{subfigure}
\caption{Visualizations of the outputs from our system. Each row visualizes RGB camera inputs, predicted road geometries, and detection and motion predictions respectively. Detection and motion prediction are used during inference; mapping is used for training only. For mapping, we predict \textbf{\textcolor[HTML]{804080}{road}},  \textbf{\textcolor[HTML]{9dea32}{broken}} and \textbf{solid} (white) lane markings. For detection, we predict pedestrians' and vehicles' poses and bounding boxes. We forecast multi-modal future trajectories with their corresponding likelihoods. Best viewed on screen.}
\lblfig{visualization}
\vspace{-1em}
\end{figure*}

\subsection{Qualitative analysis}
\reffig{visualization} provides a qualitative analysis of our system.
It shows the combined input images, LiDAR point cloud, the auxiliary segmentation predictions, detections, and predicted plans for all vehicles in the scene.
The ego-vehicles plan uses the provided high-level command, while all other vehicles predict a distribution over possible future plans.
Note how all vehicles predict a reasonable and consistent set of future plans aligning well with the inferred map-view representation of the road and the potential other vehicles.

\section{Discussion}
In this paper, we present a mapless, end-to-end driving system that trains from the experiences of all nearby vehicles.
Our system achieves state-of-the-art performance in closed-loop driving simulation, and it outperforms prior leading methods by a wide margin.
\textbf{Limitations and potential negative social impacts:}
Our approach is trained and evaluated in simulation alone and still incurs traffic infractions.
If directly deployed in the real world, it would most likely result in traffic accidents (negative social impacts).
On the technical side, our current behavior predictor instantiated by the conditional motion planner does not consider multi-modality beyond the high-level commands.
Extending our work with a probabilistic formulation will strengthen its ability in handling the diverse behaviors of both the ego vehicle and the other road users.
Improving the motion predictor beyond its raster representation is also an exciting direction.

\subsection*{Acknowlegments}
We thank Tianwei Yin for his help on pillar generation codes, Xingyi Zhou and Jeffrey Zhang for feedback on writing and figures. We thank TACC for providing part of our computing resources. This work was supported by the NSF Institute for Foundations of Machine Learning and NSF award \#1845485.

{
\bibliographystyle{ieee_fullname}
\bibliography{egbib}
}

\begin{appendices}
\section{Detailed Infractions}
In this section we report additional infraction numbers of our experiments in the main manuscript.
All infractions are measured as the number of occurrences normalized per $1$ kilometer traveled.

\subsection{Comparison with state-of-the-art}
\reftbl{sota} compares our method with baselines on the CARLA public Leaderboard~\cite{leaderboard}.
Our method also leads the red light, offroad and blocked infraction numbers among all the methods.

\begin{table*}[t]
\centering
\begin{tabular}{l@{\ }c@{\ \ \ }c@{\ }c@{\ }c@{\ }c@{\ }c@{\ }c@{\ }c@{\ }c@{\ }c@{\ }}
    \toprule
     \thead{Rank} & \thead{Method} & \thead{Driving \\ Score } & \thead{Route \\ Completion} & \thead{Infraction \\ Score} & \thead{Vehicle \\ Collisions} & \thead{Pedestrian \\ Collisions} & \thead{Layout \\ Collisions} & \thead{Red light \\ Violations} & \thead{Offroad \\ Infractions} & \thead{Blocked \\ Infractions} \\
    \cmidrule(r){1-2}
    \cmidrule(r){3-11}
    $1$ & \textbf{LAV} & $\textbf{61.85}$ & $\textbf{94.46}$ & $0.64$ & $0.70$ & $0.04$ & $\textbf{0.02}$ & $\textbf{0.17}$ & $\textbf{0.25}$ & $\textbf{0.10}$ \\
    $2$ & GRIAD & $36.79$ & $61.85$ & $0.60$ & $2.77$ & $\textbf{0.00}$ & $0.41$ & $0.48$ & $1.39$ & $0.84$ \\
    $3$ & TransFuser+ & $34.58$ & $69.84$ & $0.56$ & $0.70$ & $0.04$ & $0.03$ & $0.75$ & $0.18$ & $2.41$ \\
    $4$ & Rails~\cite{chen2021learning} & $31.37$ & $57.65$ & $0.56$ & $1.35$ & $0.61$ & $1.02$ & $0.79$ & $0.96$ & $0.47$ \\
    $5$ & IARL~\cite{toromanoff2020end} & $24.98$ & $46.97$ & $0.52$ & $2.33$ & $\textbf{0.00}$ & $2.47$ & $0.55$ & $1.82$ & $0.94$ \\
    $6$ & NEAT~\cite{Chitta2021ICCV} & $21.83$ & $41.71$ & $0.65$ & $0.74$ & $0.04$ & $0.62$ & $0.70$ & $2.68$ & $5.22$ \\
    $7$ & Transfuser~\cite{Prakash2021CVPR} & $16.93$ & $51.82$ & $0.42$ & $1.09$ & $0.91$ & $0.19$ & $1.26$ & $0.57$ & $1.96$ \\
    $8$ & LBC~\cite{chen2019lbc} & $8.94$ & $17.54$ & $\textbf{0.73}$ & $\textbf{0.40}$ & $\textbf{0.00}$ & $1.16$ & $0.71$ & $1.52$ & $4.69$ \\
    \bottomrule
\end{tabular}
\caption{Comparison of our method and the state-of-the-art on the public CARLA leaderboard~\cite{leaderboard} (accessed Jan 2022). Methods are ranked by the driving score as the main metric. Driving Score, Route Completion, Infraction Score are higher the better, whereas the rest are lower the better. Infractions are measured as number of occurences per kilometer traveled. We best all other methods by a wide margin. We significantly outperform the prior best entry by $\bf{24}$ points on the driving score, and $\bf{25}$ points on the route completion. We also lead the red light, offroad and blocked infraction numbers among all the methods.
}
\lbltbl{sota}
\end{table*}

\section{More Details}

\begin{table}[t]
\centering
\begin{tabular}{c@{\ \ \ \ \ \ }l@{\ \ \ \ \ \ }r@{\ \ \ \ }}
\toprule
\thead{Town \\ Name} & \thead{Town \\ Layout} & \thead{Number of \\ Frames}  \\
\cmidrule(r){1-1}
\cmidrule(r){2-3}
Town01 & small, EU town & $46559$ \\
Town02 & small, EU town & $63564$ \\
Town03 & large, US town & $51896$ \\
Town04 & large, US town & $46244$ \\
Town05 & large, US town & $51489$ \\
Town06 & large, US town\&highway & $41812$ \\
Town07 & small, US rural & $55465$ \\
Town10 & small, US city center & $42747$ \\
\cmidrule(r){1-1}
\cmidrule(r){2-3}
\textbf{Total} & & $\textbf{399776}$ \\
\bottomrule
\end{tabular}
\caption{Number of frames and layouts of the training towns.}
\lbltbl{dataset}
\end{table}

\begin{figure}[t]
\centering
\includegraphics[width=\linewidth]{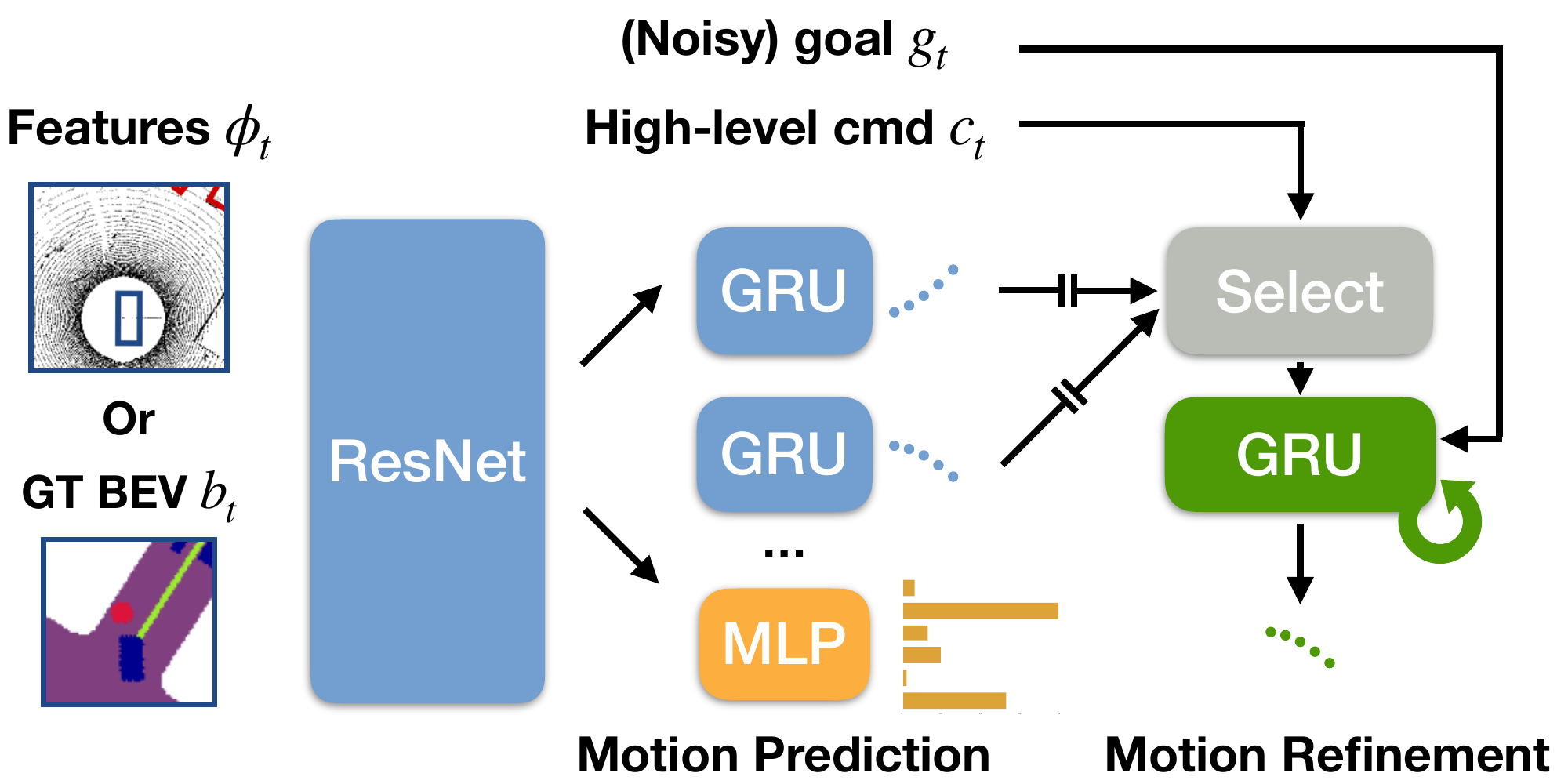}
\caption{Overview of our motion model architectures. Motion prediction outputs future trajectories on different high-level commands for all vehicles, as well as their likelihoods. For the ego-vehicle, we additionally refine it using an iterative refinement module, conditioned on $\hat{g}_t$. We detach the gradient flow from refinement to prediction in order avoid undesired causal effect.}
\lblfig{motion_detailed}
\end{figure}

\reffig{motion_detailed} provdes an overview of architectures of $M$ and $M^\prime$.
We detach the gradient of coarse trajectories predicted by $M$ to remove any effect $M_r$ might have on $M_f$ during training.

\reftbl{hyperparams} provides a list of hyperparameters. 
For all our experiments, we train our models on a 4 Titan Pascal GPU machine.

We use a ERFNet as our semantic segmentation architecture for PointPainting.
We use the following image augmentations when we train the image-based semantic segmentation and brake prediction models:
Gaussian Blur, 
Additive Gaussian Noise,
Pixel Dropout,
Multiply (scaling),
Linear Contrast,
Grayscale,
ElasticTransformation.

\subsection{Ablation study}

\reftbl{key_ablate_full} studies the effects of our key design choices.
We additionally compare to a variant where we removes PointPainting~\cite{vora2020pointpainting} (LiDAR only b.b.).
The rest of the backbone is the same.
Both Driving Score and Route Completion of this variant are lower than the full LAV.
This shows the benefit of multi-modal sensor fusion.
\reftbl{radius_ablate_full} studies the degree to which training on other vehicles' experiences affect the driving performance.
\reftbl{staged_ablate_full} studies different perception training schemes.
\reftbl{refinement_ablate_full} studies the effect of our iterative refinement module.

Note that our system does \textbf{not} rely on HD-Maps.

\subsection{CARLA Leaderboard}
Our online CARLA leaderboard entry additionally process temporal information by concatenating LiDAR scans normalized to the current vehicle pose. 
During training, this is done by stacking normalized frames using ground truth ego-vehicle poses. During testing, ego-vehicles poses are tracked across time via GNSS signals, smoothed by an Extended Kalman Filter (EKF) with a bicycle kinematics forward model.
Following prior work~\cite{toromanoff2020end}, we use a ``stuck counter'' to avoid getting blocked. We creep forward if the vehicle is stuck for too long.

\begin{table}[t]
\centering
\begin{tabular}{l@{\ }l@{\ }c@{\ }}
\toprule
\thead{Stage} & \thead{Hyperameter} & \thead{Values} \\
\midrule
\multirow{4}{*}{Privileged Motion} & batch size & 512 \\
& learning rate & 3e-4 \\
& others weight $\lambda_{other}$ & 0.5 \\
& command weight $\lambda_{cmd}$ & 0.1 \\
\midrule
\multirow{2}{*}{Perception \& Distill.} & batch size & 32 \\
& learning rate & 3e-4 \\
\bottomrule
\end{tabular}
\caption{List of hyperparameters.}
\lbltbl{hyperparams}
\end{table}

\section{Onboard Sensors}
\reftbl{sensors} provide a detailed description of the sensor configurations for the ego-vehicle.
We use the compass readings from IMU and GNSS readings to convert the target locations, represented in the GNSS format, to the ego-vehicle coordinate.

\begin{table}[t]
\centering
\begin{tabular}{l@{\ }c@{\ }l@{\ }c@{\ }}
\toprule
\thead{Count} & \thead{Modality} & \thead{Shape} & \thead{Note} \\
\midrule
$1$ & LiDAR & $\mathbb{R}^{L\times 4}$ & Velodyne-64 \\
$1$ & RGB & $\mathbb{R}^{3\times 288 \times 480}$ & FOV=$40^\circ$ \\
$3$ & RGB & $\mathbb{R}^{3\times 288 \times 256}$ & FOV=$64^\circ$ each $60^\circ$ apart \\
$1$ & IMU & $-$ &  \\
$1$ & \small{GNSS} & $\mathbb{R}^2$ & \\ 
$1$ & \small{Speedometer} & $\mathbb{R}$ & \\ 
\midrule
\end{tabular}
\caption{Configuration of our ego-vehicle's on-board sensors. The four RGB cameras are mounted at $x=1.5\text{m},y=0\text{m},z=2.4\text{m}$ with respect to the ego-vehicle's centroid.}
\lbltbl{sensors}
\end{table}

\section{Dataset Statistics}
\reftbl{dataset} describes the dataset statistics on the training towns and their corresponding layouts.
Our online leaderboard submission trains on all towns, whereas our local ablation models train on Town01, Town03, Town04 and Town06.
They test on Town02 and Town05.

\reffig{routes} provides a visualization of the four routes on which the ablation models are tested.

\begin{figure}[t]
\centering
\begin{subfigure}[b]{0.45\linewidth}
\centering
\includegraphics[width=\textwidth,trim={2cm 2cm 2cm 2cm},clip]{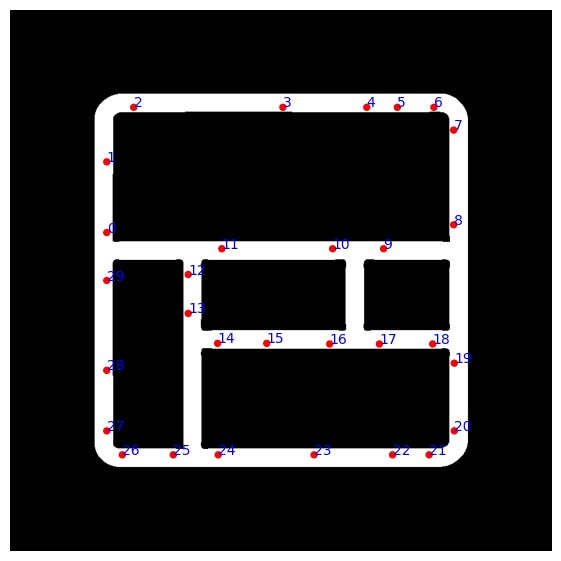}
\includegraphics[width=\textwidth,trim={2cm 2cm 2cm 2cm},clip]{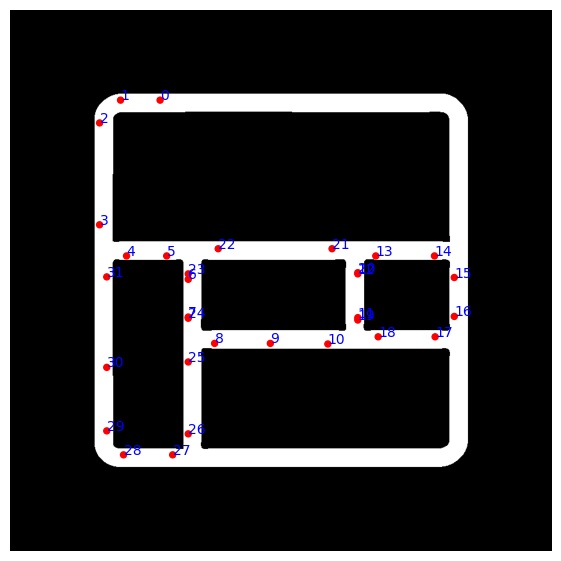}
\caption{Town02}
\end{subfigure}
\begin{subfigure}[b]{0.45\linewidth}
\centering
\includegraphics[width=\textwidth,trim={1cm 1cm 1cm 1cm},clip]{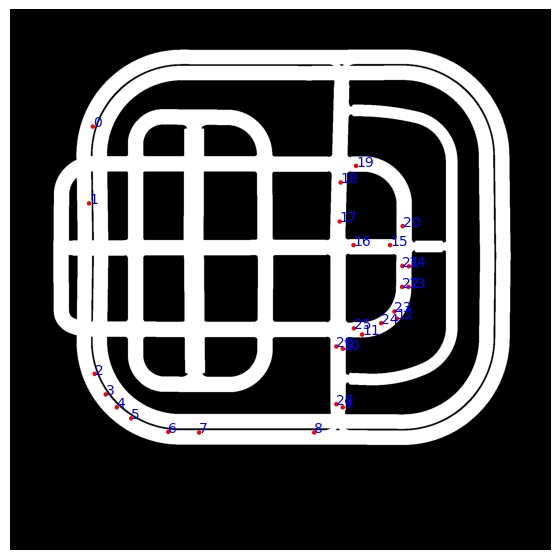}
\includegraphics[width=\textwidth,trim={1cm 1cm 1cm 1cm},clip]{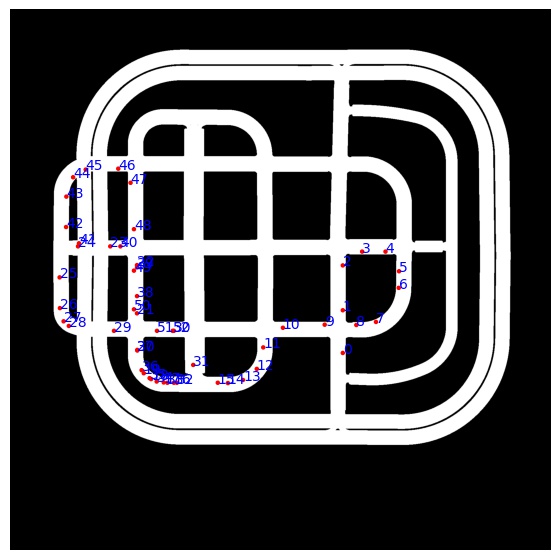}
\caption{Town05}
\end{subfigure}
\caption{Visualization of the test routes in unseen towns of the ablation models.}
\lblfig{routes}
\end{figure}

\begin{table*}[t]
\small
\centering
\begin{tabular}{l@{\ \ }c@{\ \ }c@{\ \ }c@{\ \ }c@{\ \ }c@{\ \ }c@{\ \ }c@{\ \ }c@{\ \ }c@{\ \ }}
    \toprule
     & \thead{Driving \\ Score } & \thead{Route \\ Completion} & \thead{Infraction \\ Score} & \thead{Vehicle \\ Collisions} & \thead{Pedestrian \\ Collisions} & \thead{Layout \\ Collisions} & \thead{Red light \\ Violations} & \thead{Offroad \\ Infractions} & \thead{Blocked \\ Infractions} \\
    \cmidrule(r){1-1}
    \cmidrule(r){2-10}
    \textbf{LAV} & $\textbf{45.20} \pm 6.35$ & $\textbf{91.55} \pm 5.61$ & $\textbf{0.49} \pm 0.06$ & $\textbf{0.92} \pm 0.42$ & $0.00 \pm 0.00$ & $\textbf{0.33} \pm 0.50$ & $0.28 \pm 0.28$ & $\textbf{0.27} \pm 0.01$ & $\textbf{0.01} \pm 0.02$ \\
    \midrule
    Ego-vehicle only & $38.56 \pm 1.86$ & $84.76 \pm 5.12$ & $0.46 \pm 0.02$ & $1.17 \pm 0.50$ & $0.00 \pm 0.00$ & $1.82 \pm 0.06$ & $0.34 \pm 0.20$ & $0.37 \pm 0.09$ & $0.09 \pm 0.08$ \\
    No distillation & $28.23 \pm 2.27$ & $81.05 \pm 6.04$ & $0.36 \pm 0.04$ & $2.08 \pm 0.34$ & $0.00 \pm 0.00$ & $7.87 \pm 0.15$ & $\textbf{0.21} \pm 0.04$ & $1.01 \pm 0.13$ & $0.05 \pm 0.05$ \\
    LiDAR only b.b. & $26.37 \pm 2.62$ & $74.96 \pm 4.21$ & $0.31 \pm 0.04$ & $6.51 \pm 3.03$ & $0.00 \pm 0.00$ & $0.02 \pm 0.03$ & $0.26 \pm 0.23$ & $1.56 \pm 0.71$ & $0.09 \pm 0.04$ \\
    \bottomrule
\end{tabular}
\caption{Driving performance ablation of the key components of our approach on test towns. Infractions are measured as number of occurrences per kilometer traveled. Mean and standard deviation are computed over three runs. All models are the same despite the ablated option.}
\lbltbl{key_ablate_full}
\end{table*}

\begin{table*}[t]
\centering
\begin{tabular}{l@{\ \ }c@{\ \ }c@{\ \ }c@{\ \ }c@{\ \ }c@{\ \ }c@{\ \ }c@{\ \ }c@{\ \ }c@{\ \ }}
    \toprule
    \thead{Vehicles \\ Range} & \thead{Driving \\ Score } & \thead{Route \\ Completion} & \thead{Infraction \\ Score} & \thead{Vehicle \\ Collisions} & \thead{Pedestrian \\ Collisions} & \thead{Layout \\ Collisions} & \thead{Red light \\ Violations} & \thead{Offroad \\ Infractions} & \thead{Blocked \\ Infractions} \\
    \cmidrule(r){1-1}
    \cmidrule(r){2-10}
    $\leq$ 5m  & $\textbf{46.06} \pm 1.70$ & $88.77 \pm 1.01$ & $0.51 \pm 0.02$ & $1.27 \pm 0.22$ & $0.00 \pm 0.00$ & $\textbf{0.05} \pm 0.09$ & $0.43 \pm 0.11$ & $0.69 \pm 0.09$ & $0.11 \pm 0.10$ \\
    $\leq$ 15m & $45.20 \pm 6.35$ & $\textbf{91.55} \pm 5.61$ & $0.49 \pm 0.06$ & $0.92 \pm 0.42$ & $0.00 \pm 0.00$ & $0.33 \pm 0.50$ & $0.28 \pm 0.28$ & $\textbf{0.27} \pm 0.01$ & $\textbf{0.01} \pm 0.02$ \\
    $\leq$ 25m & $37.42 \pm 3.09$ & $89.56 \pm 5.61$ & $\textbf{0.61} \pm 0.12$ & $\textbf{0.85} \pm 0.26$ & $0.00 \pm 0.00$ & $0.61 \pm 0.12$ & $\textbf{0.23} \pm 0.13$ & $0.43 \pm 0.07$ & $0.06 \pm 0.10$ \\
    \bottomrule
\end{tabular}
\caption{Driving performance in test towns of models trained with different range of other vehicles. All models are the same except for other vehicles' maximum range used during training.}
\lbltbl{radius_ablate_full}
\end{table*}

\begin{table*}[t]
\centering
\begin{tabular}{l@{\ \ }c@{\ \ }c@{\ \ }c@{\ \ }c@{\ \ }c@{\ \ }c@{\ \ }c@{\ \ }c@{\ \ }c@{\ \ }c@{\ \ }}
    \toprule
    \thead{Perception \\ Training} & \thead{Driving \\ Score } & \thead{Route \\ Completion} & \thead{Infraction \\ Score} & \thead{Vehicle \\ Collisions} & \thead{Pedestrian \\ Collisions} & \thead{Layout \\ Collisions} & \thead{Red light \\ Violations} & \thead{Offroad \\ Infractions} & \thead{Blocked \\ Infractions} \\
    \cmidrule(r){1-1}
    \cmidrule(r){2-10}
    None   & $8.47  \pm 0.83$ & $9.34  \pm 0.35$ & $\textbf{0.90} \pm 0.07$ & $2.37 \pm 2.08$ & $0.00 \pm 0.00$ & $\textbf{0.00} \pm 0.00$ & $\textbf{0.19} \pm 0.32$ & $\textbf{0.15} \pm 0.26$ & $\textbf{0.00} \pm 0.00$ \\
    Joint  & $28.36 \pm 2.11$ & $79.58 \pm 4.99$ & $0.34 \pm 0.02$ & $1.65 \pm 0.72$ & $0.00 \pm 0.00$ & $7.75 \pm 1.70$ & $0.45 \pm 0.32$ & $0.49 \pm 0.03$ & $0.24 \pm 0.21$ \\
    Staged & $\textbf{45.20} \pm 6.35$ & $\textbf{91.55} \pm 5.61$ & $0.49 \pm 0.06$ & $\textbf{0.92} \pm 0.42$ & $0.00 \pm 0.00$ & $0.33 \pm 0.50$ & $0.28 \pm 0.28$ & $0.27 \pm 0.01$ & $0.01 \pm 0.02$ \\
    \bottomrule
\end{tabular}
\caption{Driving performance in test towns of models with different perception training scheme. All models are the same except for perception training.}
\lbltbl{staged_ablate_full}
\end{table*}

\begin{table*}[t]
\centering
\begin{tabular}{l@{\ }c@{\ \ }c@{\ \ }c@{\ \ }c@{\ \ }c@{\ \ }c@{\ \ }c@{\ \ }c@{\ \ }c@{\ \ }c@{\ \ }}
    \toprule
    \thead{Refinement \\ Iteration} & \thead{Driving \\ Score } & \thead{Route \\ Completion} & \thead{Infraction \\ Score} & \thead{Vehicle \\ Collisions} & \thead{Pedestrian \\ Collisions} & \thead{Layout \\ Collisions} & \thead{Red light \\ Violations} & \thead{Offroad \\ Infractions} & \thead{Blocked \\ Infractions} \\
    \cmidrule(r){1-1}
    \cmidrule(r){2-10}
    $K=0$ & $12.69 \pm 2.86$ & $35.85 \pm 2.91$ & $0.42 \pm 0.03$ & $9.15 \pm 3.88$ & $0.00 \pm 0.00$ & $9.50 \pm 2.02$ & $0.33 \pm 0.36$ & $4.11 \pm 2.11$ & $1.41 \pm 1.22$ \\
    $K=1$ & $21.30 \pm 1.10$ & $85.90 \pm 2.46$ & $0.25 \pm 0.01$ & $2.09 \pm 0.10$ & $0.00 \pm 0.00$ & $5.58 \pm 0.28$ & $0.35 \pm 0.26$ & $0.93 \pm 0.08$ & $0.03 \pm 0.03$ \\
    $K=5$ & $\textbf{45.20} \pm 6.35$ & $\textbf{91.55} \pm 5.61$ & $\textbf{0.49} \pm 0.06$ & $\textbf{0.92} \pm 0.42$ & $0.00 \pm 0.00$ & $\textbf{0.33} \pm 0.50$ & $\textbf{0.28} \pm 0.28$ & $\textbf{0.27} \pm 0.01$ & $\textbf{0.01} \pm 0.02$ \\
    \bottomrule
\end{tabular}
\caption{Driving performance ablation on the effect of motion refinement. All models are the same except for number of refinement iterations.}
\lbltbl{refinement_ablate_full}
\end{table*}

\section{License of Assets}
We use the open source CARLA driving simulator~\cite{dosovitskiy2017carla}.
CARLA is released under the MIT license.
Its assets are under the CC-BY license.

Our teaser figure in the main paper uses a picture from the Waymo open dataset~\cite{sun2020scalability}
The Waymo open dataset uses a customized non-commercial license\footnote{\url{https://waymo.com/intl/en_us/dataset-download-terms/}}.
Part of our codebase uses the official ResNet and ERFNet implementation.
The codes are under the MIT license and the CC-BY-NC license respectively.

\end{appendices}

\end{document}